\documentclass[a4paper, 10pt, conference]{IEEEtran}
\IEEEoverridecommandlockouts
\usepackage{cite}
\usepackage{amsmath,amssymb,amsfonts}
\usepackage{algorithmic}
\usepackage{graphicx}
\usepackage{textcomp}
\usepackage{xcolor}
\def\BibTeX{{\rm B\kern-.05em{\sc i\kern-.025em b}\kern-.08em
    T\kern-.1667em\lower.7ex\hbox{E}\kern-.125emX}}
\begin{document}

\title{Continuous ErrP detections during multimodal human-robot interaction\\
\thanks{Federal Ministry for Education and Research with Grant number: 01IW21002 and Federal Ministry for Economic Affairs and Climate Action (BMWK), German Aerospace Center e.V. (DLR) with Grant number: 50RA1701, 50RA1702, 50RA1703, 50RA2023, and 50RA2024}
}

\author{Su Kyoung Kim$^{1}$, Michael Maurus$^{1}$, Mathias Trampler$^{1}$,  Marc Tabie$^{1}$, and Elsa Andrea Kirchner$^{2}$ 
\thanks{$^{1}$ S.K. Kim, Michael Maurus, Michael Maurus, and Marc Tabie are with the Robotics Innovation Center, German Research Center for Artificial Intelligence (DFKI), Robert-Hooke-Str.1, 28359, Bremen, Germany.}%
\thanks{$^{2}$ E.A. Kirchner is  with the Robotics Innovation Center, German Research Center for Artificial Intelligence (DFKI), Robert-Hooke-Str.1, 28359, Bremen, Germany and Institute of Medical Technology Systems, University of Duisburg-Essen, Bismarckstr. 8, 47507, Duisburg, Germany}%
}

\maketitle

\begin{abstract}
Human-in-the-loop approaches  are of great importance for robot applications. In the presented study, we implemented a multimodal human-robot interaction (HRI) scenario, in which a simulated robot communicates with its human partner through speech and gestures. The robot announces its intention verbally and selects the appropriate action using pointing gestures. The human partner, in turn, evaluates whether the robot's verbal announcement (intention) matches the action (pointing gesture) chosen by the robot. For cases where the verbal announcement of the robot does not match the corresponding action choice of the robot, we expect error-related potentials (ErrPs) in the human electroencephalogram (EEG). These intrinsic evaluations of robot actions by humans, evident in the EEG, were recorded in real time, continuously segmented online and classified asynchronously. For feature selection, we propose an approach that allows the combinations of forward and backward sliding windows to train a classifier. We achieved an average classification performance of 91\% across 9 subjects. 
As expected, we also observed a relatively high variability between the subjects.
In the future, the proposed feature selection approach will be extended to allow for customization of feature selection. To this end, the best combinations of forward and backward sliding windows will be automatically selected to account for inter-subject variability in classification performance. In addition, we plan to use the intrinsic human error evaluation evident in the error case by the ErrP in interactive reinforcement learning to improve multimodal human-robot interaction.
\end{abstract}

\begin{IEEEkeywords}
Multimodal human-robot interaction, brain-computer interfaces, EEG, error-related potentials, intrinsic human error evaluation
\end{IEEEkeywords}

\section{Introduction}

Electroencephalogram (EEG)-based BCIs have been well studied in various application and research areas, e.g., rehabilitation, neural engineering, robotics, etc.
In particular, multimodal interfaces that leverage intrinsic human feedback play a crucial role in improving human-robot interactions (HRI) by optimizing HRI to better interpret and thus understand the intentions of communication partners 
(see for review~\cite{kirchner2019embedded}). 
In fact, the rate of movement intentions/predictions was improved by using multimodal interfaces, which use a combination of different modalities including EEG, compared to unimodal interfaces (e.g. EEG as the only modality)~\cite{kirchner2014multimodal}.

In addition, EEG can be used to provide intrinsic feedback about the correctness of a behavior or interaction.
Error-related potentials (ErrPs) are well-known EEG components~\cite{Miltner:JCN:1997, vanSchie:NatureNeuro:1997, Falkenstein:BioPsyc:2000, Holroyd:PsychoRev:2002, Ferrez:IEEETransBioEng:2008, Iturrate:EMBS:2010, skim:smc:2013, Iturrate:ICAR:2010, skim:tnsre:2016} and research on ErrP-based BCIs has been established in various application areas (see for review~\cite{Chavarriaga:review:2014}).
In particular, ErrPs are very useful in robotics and are also well evaluated in the context of human-robot (machine) interaction (e.g.,~\cite{skim:scirep:2017, skim:icra:2020, skim:frontiers:2020}).
For example, ErrP-based BCIs were used to detect errors, and these error detections were used to correct incorrect actions of the robot without learning (e.g.,~\cite{salazar2017correcting}). The ErrP-based error detection was also used to learn behavior strategies and optimize control policy (e.g.,~\cite{Iturrate:SciR:2015, skim:scirep:2017}).
On the one hand ErrPs were used in the context of rehabilitation, in which the control policy was optimized based on ErrP classification outputs to learn the position of a robotic arm~\cite{Iturrate:SciR:2015}.
On the other hand, ErrP-based human feedback has been used for interactive reinforcement learning in robotics. 
Here, ErrPs were triggered during intrinsic error assessment in humans when different types of errors, i.e. deviations from expectation or internal models, occur and are recognized as such. This intrinsic error evaluation (ErrPs) was used for learning of human gestures in the context of human-robot interaction. In such an approach, the results of the online ErrP classification were directly used for online learning of the control strategy of a real robot~\cite{skim:scirep:2017}.

In fact, intrinsic implicit human feedback is very advantageous, since the pre-configuration of evaluation criteria (e.g., for reward shaping) does otherwise require expert knowledge.
Reward shaping is especially complicated for complex tasks or in high-dimensional workspaces. Further, ErrP-based intrinsic human feedback contains not only semantic evaluation (cf. N400~\cite{delong2020comprehending}) but also reflects subtle inconsistency (unusual situations) or subjective preference (see for review~\cite{Chavarriaga:review:2014}).

However, ErrP-based BCIs or the use of ErrP-based human feedback for robot learning is challenging because the timing of the detection  of unusual or incorrect actions is generally unknown.
ErrP detection is similarly demanding as P300 detection (synchronous classification), provided the time of a possible error is narrowly defined or the duration of the stimulus to be evaluated is very short~\cite{kirchner2016intelligent}.
However, continuous ErrP detections are necessary when the task duration is long (e.g. continuous complex behavior of robots).
Continuous detection of event-related potentials (ERP) have been investigated in movement prediction (e.g.,~\cite{Kirchner:BR:2013, kirchner2014multimodal, seeland2017adaptive}) and also in ErrP-based BCIs (e.g.,~\cite{lopes2021online}), in which sliding windows were used to continuously detect ERPs, e.g., motor-related cortical potentials (MRCPs) including lateralized readiness potential (LRP) or ErrPs.

Continuous ErrP detection requires asynchronous classifications. It is challenging to achieve a high classification performance.
We propose a feature selection approach that uses forward and backward sliding windows. These were combined in our study to find the best combination of forward and backward sliding windows to achieve optimal classification performance for continuous ErrP detections.
In this study, we show that continuous ErrP detections in multimodal human-robot interaction, where the simulated robot communicates with the human partner through speech and gestures, is possible with good classification performance.
It should be noted that we have continuously detected ErrPs online, but we have not yet shown a concrete use case of online classification results in this study, such as using ErrPs as a signal of intrinsic human evaluation of erroneous behaviour for robot learning or for simply correcting erroneous actions of the robot without learning. 
This will be future work.

\section{Methods}

\subsection{Experimental setup}

In the project TransFIT (for details, see https://robotik.dfki-bremen.de/en/research/projects/transfit/), a simulated scenario was developed for the assembly and installation of infrastructure for space applications in the context of human-robot collaboration.
Based on this scenario in a lunar environment, we have set up our experimental scenario consisting of a base camp, a workbench, solar panels, etc.
In addition, we used a simulation of our newly developed humanoid robot called RH5 Manus~\cite{2022_Boukheddimi_rh5v2}.

\begin{figure}
\vspace{0.1in}
\begin{center}
\includegraphics[width=3.3in]{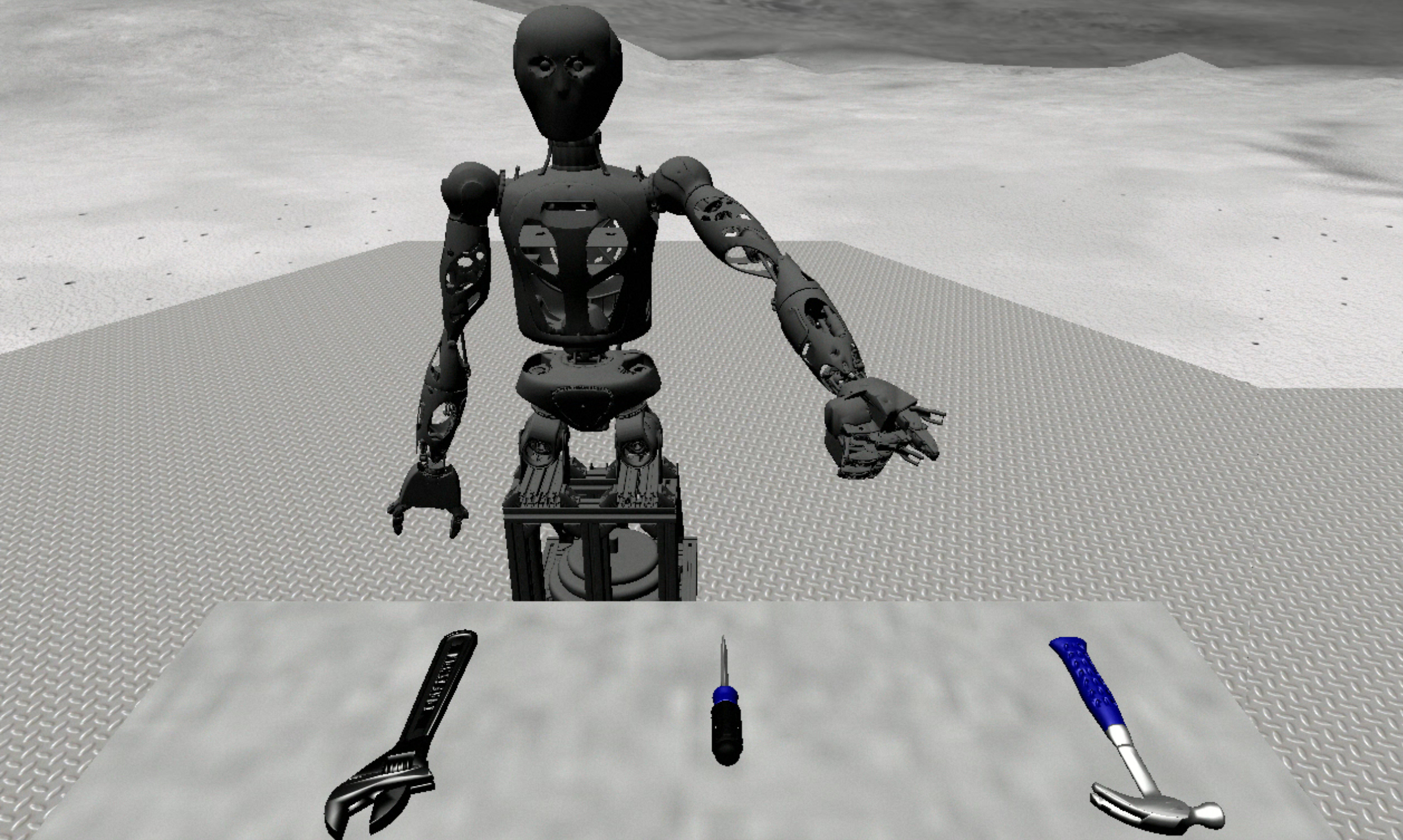}
\end{center}
\caption{Scenario: The simulated robot named RH5 Manus verbally announces which tool it intends to select and then performs a corresponding pointing action. The human partner in turn evaluates whether the robot's verbal announcement matches the robot's action. For example, if the robot verbally announces that the hammer will be selected and then points to the hammer, the robot's action is correct. In this case, we do not expect any error-related potentials. However, if the robot's verbal announcement does not match the robot's expected matching action, ErrPs will be evoked during the execution of the robot's pointing gesture.}
\label{fig:sce}
\end{figure}

In our scenario shown in Figure~\ref{fig:sce}, the simulated RH5 Manus is able to give verbal information about intended behaviour. 
It can also point to objects in the simulation.
In our experiments, RH5 Manus first verbally announces which object he will point to next, and then performs the action either correctly or incorrectly by pointing to different objects.
If the mapping between the verbal announcements and the corresponding pointing gestures is incorrect,
we expect error-related potentials (ErrPs) to be elicited in the subject's brain and measured in the EEG.
In the simulation, three different objects were available on the workbench: a spanner, a screwdriver and a hammer.

\begin{figure}
\vspace{0.1in}
\begin{center}
\includegraphics[width=3.3in]{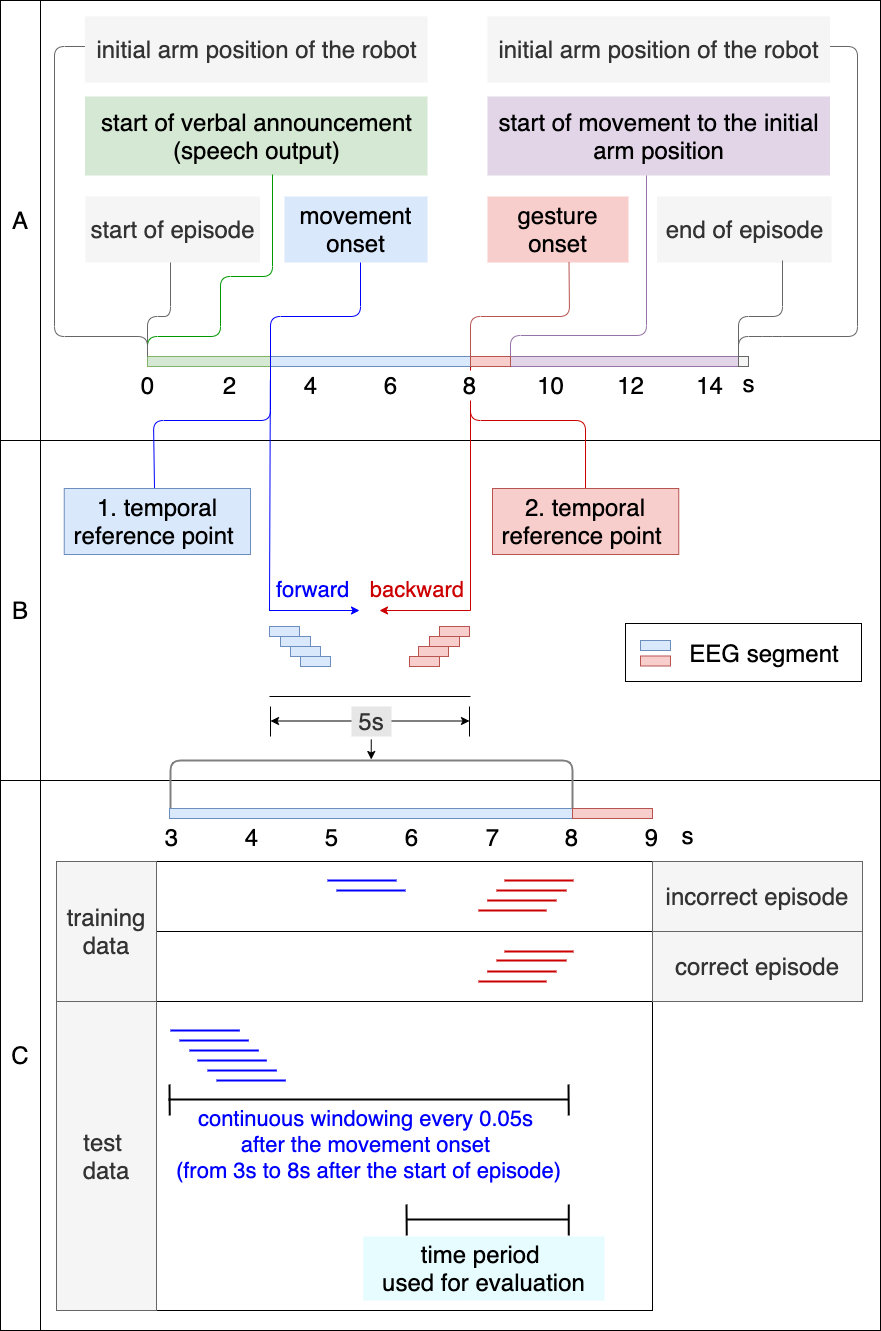}
\end{center}
\caption{Experiment design. (A) Episode: An episode begins with the start of the robot's verbal announcement and ends with the return to 
the initial position. (B) Concept of forward and backward sliding windows used for training a classifier: Features are extracted from the time period between the onset of movement and the onset of gesture using forward and backward sliding windows. Note that we divided the robot's action into different phases (directional movements and gesture movements), but the robot performs a continuous action to point to one of three objects. (C) Feature selection during training for correct and incorrect episodes and during continuous testing. Evaluation is based on the marked time period.}
\label{fig:exp_design}
\end{figure}

Figure~\ref{fig:exp_design}-A shows our experimental design.
In the experiments, we defined episodes, each lasting from the beginning of the robot's verbal announcement until the end of the robot's action (see Fig.~\ref{fig:exp_design}-A).
Each action of the robot consists of a movement of the arm in the direction of the workbench (here called movement) and a subsequent gesture lasting approximately 1 second (here called gesture) indicated by the change in finger configuration.
We have divided all episodes into two different types: correct and incorrect episodes.
Both labels (correct and incorrect episodes) were used to train an ErrP decoder (supervised learning) and later to validate the test data (ground truth).
We did not expect ErrPs to occur in correct episodes, while ErrPs were expected in wrong episodes.
In addition, the beginning of the robot's actions (directional movements to point at an object) and
the beginning of the robot's gestures (change of finger configuration in the robot hand for a pointing gesture) were used as temporal reference points (see Fig. ~\ref{fig:exp_design}-B), which were continuously sent as markers to the EEG recording system.
In addition, the beginning of the episode (i.e. the beginning of the verbal announcement) was written into the EEG as another marker.

Figure~\ref{fig:exp_design}-B shows our approach of feature selection/extraction using forward and backward sliding windows.
In our scenario, we do not know the exact moment when the subject realizes that the robot's actions may or may not be correct.
Therefore, we detect ErrPs asynchronously.
Two points in time are relevant for the detection of ErrPs: the beginning of the robot's actions (directional movements)
and the beginning of pointing gestures (indicated by the change in finger configuration).
First, the subject recognizes the robot's direction of movement and may tend to guess which tool the robot might select early after the start of an action (directional movements).
However, the subject cannot be absolutely sure that the robot's action is correct
until the robot performs a pointing gesture (second time point).
Therefore, we defined two temporal reference points (see Fig.~\ref{fig:exp_design}): (a) directional movements towards one of the objects ($3$s-$8$s after verbal announcement) and (b) pointing gestures towards the selected object ($8$s-$9$s after verbal announcement) to frame the time period during which the subject makes a decision about the correctness of the robot's action.
For training an ErrP decoder, we used these two temporal reference points for feature selection and extraction.
As shown in Figure~\ref{fig:exp_design}-B, on the one hand, we continuously segmented the EEGs from the onset of the robot actions (i.e. forward windowing).
On the other hand, we continuously segmented from the onset of the robot's gestures in the reverse direction (i.e., backward windowing).
In this way, features are extracted using forward and backward sliding windows (for details, see section~\ref{EEG methods}).

\subsection{Dataset}
Nine subjects ($2$ females, $7$ males, age: $25.5\pm3.02$ years,
right-handed, normal or corrected-to normal vision)
participated in this study.
The experimental protocols were approved by the ethics committee of the University of Bremen.
Written informed consent was obtained from all participants.

For each subject we recorded 9 data sets. For one of the subjects, only 8 datasets were recorded.
Each dataset contained $36$ correct and $18$ wrong episodes.
In total, we recorded $324$ correct episodes and $162$ wrong episodes for each subject except for one subject.
Since we only had 8 complete data sets for all 9 subjects, 
7 data sets were used to train an ErrP decoder and one dataset was used for testing to allow a fair comparison between subjects.

\subsection{EEG recording, preprocessing and classification}
\label{EEG methods}

\begin{figure}
\vspace{0.1in}
\begin{center}
\includegraphics[width=3.3in]{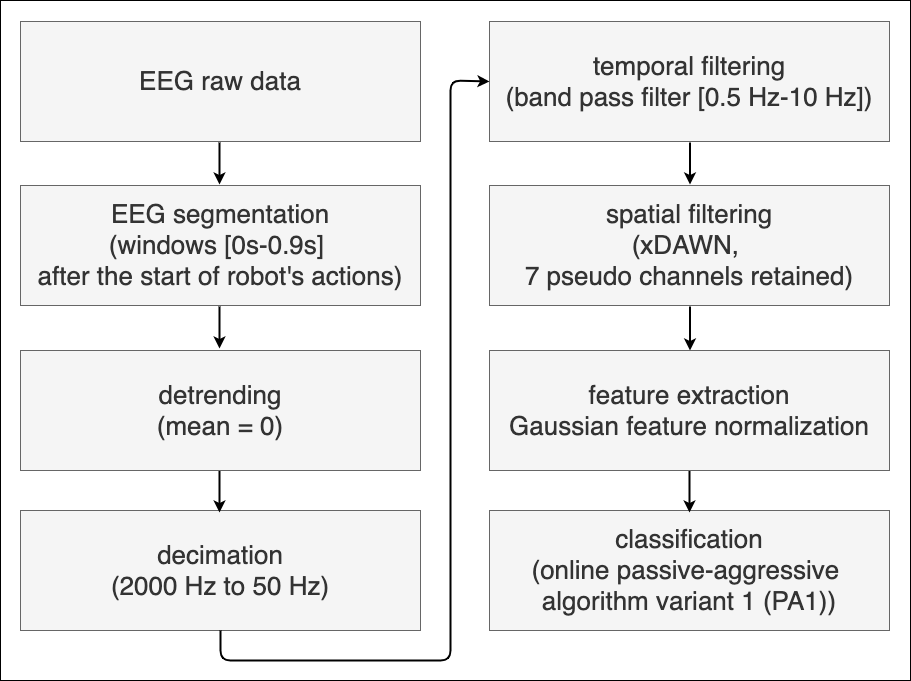}
\end{center}
\caption{EEG preprocessing and classification:
EEGs were segmented, normalized, decimated, and bandpass filtered. A spatial filter called xDAWN~\cite{Rivet:IEEE_TBE:2009}
was applied to enhance the signal-to-noise ratio and to reduce the dimensionality. Features were extracted from seven pseudo channels.
Details of feature selection and extraction are shown in Fig.~\ref{fig:exp_design}.
The online passive-aggressive algorithms variant 1 (PA1)~\cite{crammer_online_2006} was used for classification.}
\label{fig:eeg_processing}
\end{figure}

EEGs were continuously recorded using a $64$-channel eego mylab system (ANT Neuro GmbH),
in which 64 electrodes were arranged in accordance to an extended $10$-$20$ system
with reference at electrode CPz. Impedance was kept below $10$\,k$\Omega$.
EEG signals were sampled at $2$\,kHz amplified by one $64$ channel amplifiers (ANT Neuro GmbH).

Figure~\ref{fig:eeg_processing} shows the data flow for EEG analysis.
The EEG data were analyzed using a Python-based framework
for signal processing and classification~\cite{Krell:pySPACE:2013}.
The continuous EEG signal was segmented into epochs from $0$\,s to $0.9$\,s
after the start of the robot's action with the overlap of $0.05$\,s (sliding windows).
All epochs were normalized to zero mean for each channel, decimated to $50$\,Hz,
and band pass filtered ($0.5$ to $10$\,Hz).
The xDAWN spatial filter~\cite{Rivet:IEEE_TBE:2009} was used to enhance
the signal to signal-plus-noise ratio and $7$ pseudo channels were retained after spatial filtering.

For feature selection and extraction,
we divided the robot's action into different phases (directional movements and gesture movements).
However, we would like to point out that the robot performs a continuous action to point at one of the three objects.
As mentioned earlier, the human observer can recognize the direction of movement and guess the robot's choice at the beginning of the robot's action (directional movements).
However, the human observer cannot be absolutely sure that the robot's actions are correct until the robot performs a pointing gesture (gesture movements).
Two temporal reference points were relevant for feature selection and extraction:
The onset of the robot's action (i.e., the onset of directional movements) and the onset of gesture movements (see Fig.~\ref{fig:exp_design}-B).
Accordingly, we extracted features using forward and backward windowing based on the two temporal reference points (see Fig.~\ref{fig:exp_design}-B).

Different strategies of feature selection and extraction were used depending on the episode type after a systematic investigation of different combinations of sliding windows across both types of episodes for each subject. We found different optimal time periods and different optimal combinations of forward and backward sliding windows for feature selection and extraction depending on the individual subject. However, to allow a fair comparison between subjects, we decided to choose the same time period for all subjects. Therefore, we used the same sliding windows and the same combinations of forward and backward sliding windows for all subjects to train a classifier.

The final selected time period for correct and incorrect episodes as follows.
For incorrect episodes, four windows of $0.9$s in length were used, ending at $0$s, $-0.05$s, $-0.1$s and $-0.15$s with respect to the beginning of the gesture movement (the second temporal reference point). That is, four windows were segmented from the beginning of the gesture movements in the reverse direction. In addition, two windows were used with a length of $0.9$s from $2$s and $2.5$s with respect to the beginning of the directional movement (the first temporal reference point).
For correct episodes, four windows of $0.9$s length were likewise used, ending at $0$s, $-0.05$s, $-0.1$s and $-0.15$s with respect to the onset of gesture movement (the second temporal reference point). However, we did not use the windows that can be segmented from the first temporal reference point (the beginning of the directional movement). 

For testing, ErrPs are continuously detected every $0.05$s with a window length of $0.9$s from the start of the robot action.
This means that ErrPs were continuously detected in the time period between the start of the robot's action and the end of the robot's pointing gesture.
For evaluation, we used the time period between $6$s and $8$s after the start of the episode (i.e., from $3$s to $5$s after the movement onset) as the ground truth (see Fig. ~\ref{fig:exp_design}-C). This means that the sliding windows that end in this time period were used to evaluate the trained classifier. During this time period, we can ensure that the robot's arm position is unambiguous for the subjects' evaluation.

The features were normalized and used to train a classifier.
The online passive aggressive algorithm variant 1 (PA1) ~\cite{crammer_online_2006} was used as classifier.
The cost parameter of the PA1 was optimized using a grid search,
in which an internal stratified 5 fold cross validation was performed on the training data (7 training datasets)
and the best value of [$10^{0}$, $10^{-1}$, ... , $10^{-6}$] was selected.
The performance metric used was balanced accuracy (bACC), which is an arithmetic mean of
true positive rate (TPR) and true negative rate (TNR).

\section{Results and Discussion}
Table~\ref{tab:cl} shows the classification performance for each subject. We achieved an average classification performance of $91$$\%$ across all subjects. However, we found a high variability between subjects.

\begin{table}[htbp]
\renewcommand{\arraystretch}{1.3}
\caption{Classification performance}
\begin{center}
\begin{tabular}{|c|c|}
\hline
subjects  & balanced accuracy (bACC)$^{\mathrm{a}}$\\
\hline
subject 1 & 0.89\\
subject 2 & 0.94\\
subject 3 & 0.97\\
subject 4 & 0.83\\
subject 5 & 0.94\\
subject 6 & 0.92\\
subject 7 & 0.97\\
subject 8 & 0.90\\
subject 9 & 0.80\\
\hline
mean and standard deviation & 0.91$\pm$0.06\\
\hline
\multicolumn{2}{l}{$^{\mathrm{a}}$ bACC = (TPR+TNR)/2}
\end{tabular}
\label{tab:cl}
\end{center}
\end{table}

The inter-subject variability is not surprising, as subjects have different strategies to evaluate the continuous actions of the robot.
The duration of the task was not short ($5$s from the beginning of the robot action to the start of the robot pointing gestures).
Therefore, we assume that each subject evaluates the correctness of the robot's choice of action at different points in time.
One can estimate the direction of the robot's movement immediately after the robot starts moving and focus less on the robot's pointing gestures.
In this case, the pointing gestures can even be overlooked unintentionally.
Another person may focus only on the robot's pointing gestures.
Of course, some subjects also focus on the overall action of the robot.
In particular, jerky movements of the robot can affect the accuracy of the correct estimation of the robot's direction of movement. In fact, some subjects reported that jerky movements affect the estimation of the robot's direction of movement.
Furthermore, the evaluation by the subject can be changed during the execution of the action by the robot.
For example, the robot's action may initially be evaluated as a correct action after the direction of the arm movement has been detected, 
but this evaluation may be changed with the execution of the pointing gesture.

For this reason, we proposed an approach using forward and backward sliding windows and their combinations.
Nevertheless, we did not achieve high classification performance for some subjects.
The reason for this could be that we used the same combination of forward and backward sliding windows for all subjects.
As mentioned earlier, we investigated different combinations of sliding windows across both types of episodes for each subject.
We found different optimal time periods and the different optimal combinations of forward and backward sliding windows across both episode types for each subject.
On the one hand, this means that the optimal time periods relevant for recognizing the correct actions differ from subject to subject.
On the other hand, the optimal time periods that are relevant for detecting the wrong actions also vary from subject to subject.
Furthermore, the combination of the optimal time periods for correct and incorrect actions also varies between subjects.
We nevertheless chose to apply the same combination of forward and backward sliding windows to all subjects to allow a fair comparison between subjects.

Therefore, in the future, it makes sense to extend the proposed approach of forward and backward sliding windows and their combinations with an individual adjustment of feature selection so that high inter-subject variability in classification performance is avoided.
With such an individual adjustment of feature selection, classification performance can be improved if we identify the best feature combinations for each subject.
However, such investigations are time-consuming. Therefore, developing an automatic selection of best combinations to customize the feature selection is useful.
In the future, it is useful to extend the proposed approach of feature selection using forward and backward sliding windows and their combinations for individual adaptions of feature selection, in which the best individual combinations are automatically computed depending on different context, e.g., different scenarios and different situations.

Furthermore, although we continuously detected ErrPs online, we did not use the results of online classification for intrinsic online corrections of the robot's erroneous actions or for robot learning (e.g., optimizing learning strategies based on ErrP-based human feedback) in this scenario, such as in our previous studies~\cite{skim:scirep:2017, skim:icra:2020, skim:frontiers:2020}. In this previous work, we have used ErrP-based human feedback online for intrinsic interactive reinforcement learning in real human-robot interaction to optimize the robot's learning strategy in real time, but ErrPs did not need to be classified asynchronously due to the short task duration of the robot. Therefore, in a next step, we plan to use ErrP-based human error evaluation for intrinsic interactive reinforcement learning, where continuous ErrP detections are necessary during continuous and long-lasting task execution of a robot (e.g. continuous complex robot behavior) and we therefore expect e.g., more than one type of error during the execution of a task.

\bibliographystyle{IEEEtran}
\bibliography{references}

\end{document}